%% file: book.tex
\documentclass[openany,ngerman]{book}

\usepackage[english]{babel}
\usepackage[utf8]{inputenc}%
\usepackage[small]{dgruyter}
\usepackage{microtype}
\usepackage{makeidx}
\makeindex

\include{includes/includes_kiel_university}

\author{H\"{u}seyin Abut, Gerhard Schmidt, Kazuya Takeda, Jacob Lambert, and John H.L. Hansen}
\title{Intelligent Vehicles and Transportation}

\distributionseries{Towards Human-Vehicle Harmonization}
\seriestitle{Intelligent Vehicles and Transportation}
\seriessubtitle{Towards Human-Vehicle Harmonization}
\serieseditor{H\"{u}seyin Abut, Gerhard Schmidt, Kazuya Takeda, Jacob Lambert, and John H.L. Hansen}
\seriesvolume{3/5}

\subtitle{Towards Human-Vehicle Harmonization}

\copyrighttext{Copyright-Text}
\isbn{978-3-11-021808-4}
\eisbnpdf{978-3-11-021809-1}
\issn{0179-0986}
\cover{Cover-Firma}
\typesetter{le-tex publishing services GmbH, Leipzig}
\printbind{Druckerei XYZ}

\input{content/preamble}

\sidecaptionvpos{figure}{c}

\newacro{CADC}{Canadian Adverse Driving Conditions}
\newacro{WADS}{Winter Adverse Driving dataSet}
\newacro{ACDC}{Adverse Conditions Dataset with Correspondences}

\newacro{SOR}{Statistical Outlier Removal}
\newacro{ROR} {Radius Outlier Removal}
\newacro{DROR}{Dynamic Radius Outlier Removal}
\newacro{DSOR}{Dynamic Statistical Outlier Removal}
\newacro{FCSOR}{Fast Cluster Statistical Outlier Removal}
\newacro{LIOR}{Low Intensity Outlier Removal}
\newacro{DIOR}{Dynamic light-Intensity Outlier Removal}
\newacro{DDIOR}{Dynamic Distance-Intensity Outlier Removal}

\newacro{ICP}{Iterative Closest Point}
\newacro{IMU}{Inertial Measurement Unit}
\newacro{FPGA}{Field-Programmable Gate Array}
\newacro{SLAM}{Simultaneous localization and mapping}
\newacro{RPE}{Relative Pose Error}
\newacro{lidar}{Light Detection and Ranging}
\newacro{GNSS}{Global Navigation Satellite System}
\newacro{RTK}{Real Time Kinematic}
\newacro{Cerema}{Centre for Studies and Expertise on Risks, the Environment, Mobility and Urban Planning}

\begin{document}

\frontmatter

\include{content/main}

\backmatter

\end{document}

%% file: includes/includes_kiel_university.tex
\usepackage{psfrag}		
\usepackage{booktabs}
\usepackage{rotating}
\usepackage{epsfig}
\usepackage{makeidx}

%% file: content/preamble.tex
\usepackage[l2tabu,orthodox]{nag}

\usepackage[
    backend=bibtex8,
    style=vancouver,
    sorting=none,
    natbib=true,
    doi=false,
    isbn=false,
    url=false,
    eprint=false,
    maxcitenames=1,
    mincitenames=1
]{biblatex}

\DeclareFieldFormat{labelnumberwidth}{\textrm{[#1]}\midsentence}
\DeclareNameWrapperFormat{author}{\textrm{#1}}
\DeclareFieldFormat[article]{title}{\textrm{#1}}
\DeclareFieldFormat[article]{journaltitle}{\textrm{#1}.}
\DeclareFieldFormat[article]{volume}{\textrm{#1\iffieldundef{number}{}{(\printfield{number})}:}}
\DeclareFieldFormat[article]{pages}{\iffieldundef{volume}{:}{}\textrm{#1}}
\DeclareFieldFormat[article]{date}{\textrm{#1}}

\DeclareFieldFormat[inproceedings]{title}{\textrm{#1}}
\DeclareFieldFormat[inproceedings]{booktitle}{\textrm{#1}}
\DeclareFieldFormat[inproceedings]{address}{\textrm{#1}}
\DeclareFieldFormat[inproceedings]{pages}{\iffieldundef{volume}{;}{}\textrm{#1}}
\DeclareFieldFormat[inproceedings]{date}{\textrm{#1}}
\DeclareListFormat[inproceedings]{location}{\textrm{#1}}
\DeclareListFormat[inproceedings]{publisher}{\textrm{#1}}
\DeclareNameWrapperFormat{location}{\textbf{#1}}
\DeclareFieldFormat[inproceedings]{publisher}{\textrm{#1}}

\addto\extrasenglish{
    
}

\usepackage[printonlyused]{acronym}

\usepackage{siunitx}
\usepackage[all]{nowidow}

\usepackage[dvipsnames]{xcolor}

\usepackage{lipsum}

\usepackage{xspace} %
\newcommand{\ie}{i.e., \xspace{}}

\usepackage{subcaption}

\usepackage{import}

\usepackage{booktabs}

\usepackage{pifont}
\usepackage{tabularx}
\usepackage{multirow, multicol}

\usepackage{amssymb,amsfonts,amsmath,amscd}

\usepackage{bm}

\newcommand{\bbm}{\begin{bmatrix}}
\newcommand{\ebm}{\end{bmatrix}}

\usepackage{commath}

\newcommand{\SR}{\mathbb{R}}
\renewcommand{\P}[1]{\mathbb{P}\left(#1\right)}
\newcommand{\expp}[1]{\exp\left(#1\right)}

%% file: content/main.tex
\contribution

\contributionauthor{Cl\'{e}ment Courcelle, Dominic Baril, Fran\c{c}ois Pomerleau and Johann Laconte}
\affil{Northern Robotics Laboratory, Universit\'{e} Laval,
Qu\'{e}bec, QC, Canada G1V 0A6}
\runningauthor{Cl\'{e}ment Courcelle et al.}
\contributiontitle{On the Importance of Quantifying Visibility for Autonomous Vehicles under Extreme Precipitation}
\runningtitle{Quantifying Visibility for Autonomous Vehicles under Extreme Precipitation}

\input{content/abstract}

\makecontributiontitle

\input{content/intro}

\input{content/rw}

\input{content/theory}

\input{content/results}

\input{content/conclusion}

\begingroup
\let\clearpage\relax
\printbibliography
\endgroup

%% file: content/abstract.tex
\abstract{
In the context of autonomous driving, vehicles are inherently bound to encounter more extreme weather during which public safety must be ensured.
As climate is quickly changing, the frequency of heavy snowstorms is expected to increase and become a major threat to safe navigation.
While there is much literature aiming to improve navigation resiliency to winter conditions, there is a lack of standard metrics to quantify the loss of visibility of lidar sensors related to precipitation.
This chapter proposes a novel metric to quantify the lidar visibility loss in real time, relying on the notion of visibility from the meteorology research field.
We evaluate this metric on the \ac{CADC} dataset, correlate it with the performance of a state-of-the-art lidar-based localization algorithm, and evaluate the benefit of filtering point clouds before the localization process.
We show that the \ac{ICP} algorithm is surprisingly robust against snowfalls, but abrupt events, such as snow gusts, can greatly hinder its accuracy.
We discuss such events and demonstrate the need for better datasets focusing on these extreme events to quantify their effect.
}

\keywords{Localization, Snow, Precipitation, Visibility, Lidar}

%% file: content/intro.tex
\section{Introduction}
\label{CBPL_Sec:Introduction}

\begin{SCfigure}
	\includegraphics[width=.65\linewidth]{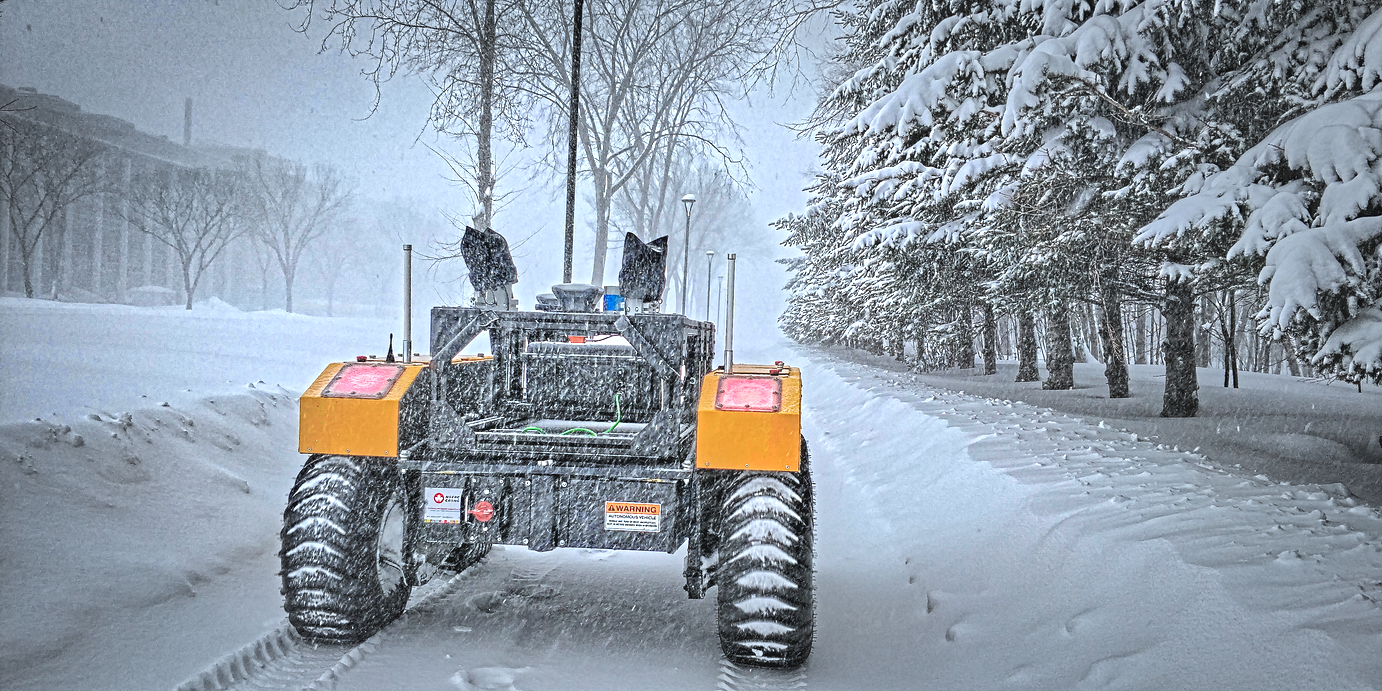}
    \caption{Robot driving in a snowstorm in Quebec City, Canada. 
    Harsh snowstorms yield considerable noise on lidar data, especially during snow gusts, potentially leading to localization failure.} 
	\label{CBPL_fig:intro}
\end{SCfigure}

As robotics becomes more and more part of our everyday life, autonomous vehicles start to roam the world alongside human drivers.
In this regard, localization is a crucial component of autonomous navigation.
Lidar \index{lidar}sensors have emerged as a standard for autonomous vehicle sensor suites, providing dense information about the environment~\cite{Roriz2021}.
However, lidars can be sensitive to the noise created by adverse weather like rain, fog or snow because of the reflection of the laser beams by raindrops or snowflakes~\cite{Rasshofer2011}.
\autoref{CBPL_fig:intro} shows an example of heavy snowfall that happened during the winter in the campus of Quebec City, Canada.
Such a level of precipitation yields consequent noise on lidar data, where most of the lidar beams were stopped by a snowflake, thus returning erroneous data.
To this effect, various snow removal \index{filters}filters have recently been proposed in the literature for lidar scans~\cite{Charron2018, Kurup2021}.
Datasets have also been released with the aim of developing and evaluating localization algorithms robust to precipitation, such as the \acf{CADC} dataset~\cite{Pitropov2021}.
However, most work in the literature uses qualitative metrics and there is no standard protocol to quantify the snow precipitation in the vicinity of the vehicle.

In this chapter, we propose a novel method to physically model the noise encountered by lidars in \index{snowstorms}snowstorms.
Using this method, we are able to quantify the level of precipitation using only lidar information, relying on the notion of visibility from the meteorology research field.
We then evaluate this metric on all sequences from the \ac{CADC} dataset and correlate it with the localization performance of a state-of-the-art lidar-based localization algorithm.
We demonstrate that using the data available in the literature, lidar-based localization is currently robust to significant snowfall.
Nevertheless, we observe that localization algorithms will potentially fail under extreme snowfall.
This observation motivates additional datasets that better feature extreme weather, effectively pushing localization algorithms to their limit.
Therefore, the contributions of this chapter are 
1) a novel metric to estimate the visibility in snowstorms; and 
2) an in-depth analysis of the impact of snowstorms on lidar-based localization.

%% file: content/rw.tex
\section{Related Work}
\label{CBPL_Sec:related_work}

Measurements of \index{precipitation}precipitation type and intensity are often done through the work of hydrologists and meteorologists through the use of a disdrometer device~\citep{Tokay2014}.
These devices provide precipitation rates based on the measures of the size and the velocity of precipitation particles.
\citet{Filgueira2017} measured the rain intensity by using the rates provided by a meteorological station to quantify the performance of lidar sensors under rain.
For a similar objective in rain and fog, \citet{Kutila2018} characterized weather by measuring meteorological visibility, particle size distribution, and rainfall intensity.
The evaluations were performed indoor in fog chamber facilities, allowing the use of several weather sensors. 
To provide outdoor field measurements, various robotic deployments and datasets under snow precipitation have recently been documented in the literature, measuring either the intensity or the density of precipitation.
\citet{MacTavish2018} have deployed a vision-based teach-and-repeat navigation system for 100 days at the University of Toronto Institute for Aerospace Studies, conducting multi-seasonal, off-road navigation, for a total of over~\SI{29}{\kilo\meter}.
Unfortunately, the intensity of snowfalls is only qualitatively described in this work.
Later, \citet{Baril2022} deployed a similar teach-and-repeat off-road navigation framework, but this time based on lidars. 
The authors cumulated~\SI{18.8}{\kilo\meter} of autonomous navigation data in boreal forest and classified the precipitation using the disdrometer measures of a nearby meteorological station.
\citet{Pitropov2021} have released the \ac{CADC} multi-modal dataset including over~\SI{20}{\kilo\meter} of driving through two years in winter under harsh weather in Waterloo, Canada.
For this dataset, the snow density is estimated by the number of points that are removed by the \ac{DROR} filter~\citep{Charron2018}.
This metric is dependent on the lidar sensor used for the dataset, which is a Velodyne VLP-32C. %
Alternatively, \citet{Kurup2021} have released the \ac{WADS}, recorded in the Keweenaw peninsula in Michigan. 
In this dataset, the authors quantify the intensity of snowfalls by measuring the snowfall rate in inches per hour.
Lastly,~\citet{Burnett2022Boreas} have proposed the Boreas dataset, featuring over~\SI{350}{\kilo\meter} of urban driving data in Canadian winter.
Precipitation characterization is limited to rain or snow in the Boreal dataset, indicating neither its intensity nor its density.
In this work, we propose a novel metric to estimate the visibility in precipitation that is adapted for real-time computation and mobile sensors measurements, while correlating with the field of meteorology. 
Our goal is that this metric allows standardizing how precipitation is quantified in various datasets while being computable in real time for autonomous platforms. 

Driving in \index{adverse weather}adverse weather has revealed to be a major challenge in autonomous navigation~\cite{VanBrummelen2018}, leading to a lot of research aiming at analyzing and reducing the impact of precipitation on different aspects of autonomous navigation, such as localization and object detection.
Moreover, \citet{Burnett2022} showed that lidar-based mapping and localization algorithms perform surprisingly well in snowstorms and even exceed radar-based algorithms.
This result was also observed in deployment by \citet{Baril2022} who noted that it is not the snowfalls but rather snow accumulated on the ground that disrupts the mapping and localization.
Various filters have also been proposed in the literature to remove the precipitation noise on lidar scans.
Firstly, the \ac{ROR} filter is included in the Point Cloud Library~\cite{Rusu2011}.
It is not designed for snow specifically and therefore does not take into account that the snow-induced noise is stronger close to the sensor, as shown by \citet{Michaud2015}.
To address this limitation, \citet{Charron2018} developed the \ac{DROR} filter. 
This filter is based on the \ac{ROR}, but the searching radius depends on the distance between the current point and the sensor.
This modification greatly improves the filter efficiency for snow points further than five meters away from the sensor.
Secondly, the \ac{SOR} filter is also implemented in the Point Cloud Library~\cite{Rusu2011}. 
Similar to the \ac{ROR} filter, it was not designed to remove snow-induced noise specifically.
Thus, \citet{Kurup2021} designed the \ac{DSOR} filter, using a threshold proportional to the distance between the point and the sensor. 
The \ac{DSOR} filter achieves better recall but a slightly lower precision than the \ac{DROR}, meaning that \ac{DSOR} removes more snow but also more environmental points. 
These works do not evaluate the improvement of localization after the filter application, which was first done by \citet{Wang2022}, who present the \ac{DDIOR} filter, relying on point intensity and neighborhood.
The evaluation of the localization accuracy was performed on one sequence of the \ac{WADS} dataset.
In this work, we correlate the accuracy of a state-of-the-art localization algorithm with snowstorm intensity.
This snowstorm intensity is computed using the novel visibility metric that we present in the next section.

%% file: content/theory.tex
\section{Theory}
\label{CBPL_Sec:Theory}

In this section, we look at the development of a density measure based on the work of \citet{Laconte2019} who developed a method to measure \index{density fields}density fields in the case of obstacle avoidance for safe navigation.
We propose to extend their method to our work, supplanting the obstacles field by the density field of the snowstorm.
This allows us to define the notion of \index{visibility}visibility in the context of lidar measurements, with a strong connection with the notion of visibility in the field of \index{meteorology}meteorology \cite{Middleton1957, Rasmussen1999}.
In the following, we propose a method to estimate the snow density field around the robot.
We assume that the density of the snow can be modeled using a heterogeneous Poisson point process, and that the density of the snow is constant in the $z$ direction for a small height difference.
As such, we only take into account points that fall within a \SI{1}{\m} strip in the $z$ direction, centered on the position of the lidar. 
The probability $\P{\mathtt{object}}$ that a beam goes through a space $\mathcal{S}\subset\SR^2$ and measures an object without colliding with a snowflake can be computed as
\begin{equation}
  \begin{aligned}
    \P{\mathtt{object}} &= \expp{-\Lambda(\mathcal{S})}, \quad\text{with } \Lambda(\mathcal{S}) = \int_\mathcal{S} \lambda(\bm{s})\dif \bm{s},
  \end{aligned}
\end{equation}
where $\lambda(\bm{s})\in\SR_{>0}$ is the density of snowflakes at the position $\bm{s}\in\SR^2$.

To estimate and store the density field, the space is tessellated into cells.
As such, each cell $c_i$ contains a density $\lambda_i$ corresponding to the associated density at this position.
Moreover, the lidar sensor is modeled as having a collision area $A_c$, meaning that for each measurement, the snowflake that caused the beam to be reflected was in a region of area $A_c$ around the measurement.
As a recall, the density field can be estimated with
\begin{equation}
  \lambda_i = \frac{1}{A_c}\ln\left(1+\frac{h_i}{m_i}\right),
  \label{eq:lambda}
\end{equation}
where $h_i$ is the number of times a collision occurred in the cell $c_i$ and $m_i$ the number of times a beam went through the cell $c_i$ without collision~\cite{Laconte2019}.
For extensive details on the mathematical proofs, please refer to~\cite{Laconte2021}.

\begin{figure*}
        \includegraphics[width=\linewidth]{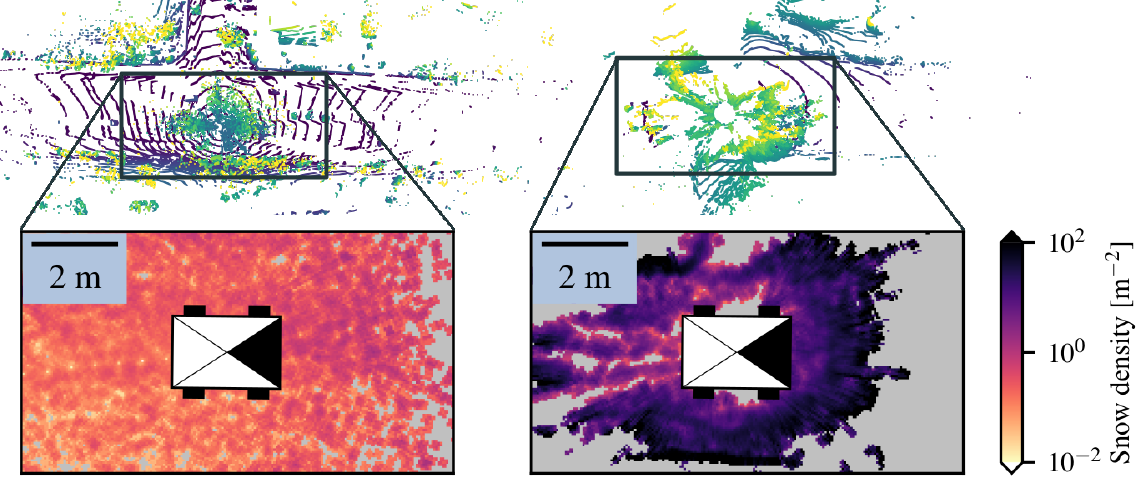}
        \caption{
        Examples of snow density fields with their associated point clouds.
        Point clouds are colored according to the point's height for easier reading.
        In the zoomed rectangles, gray areas correspond to unobserved space by the lidar beams.
        Left: The robot is navigating while a weak snowstorm is in effect, resulting in a density field of low values.
        Most of the environment is still visible, where walls and other landmarks are easily recognizable.
        Right: The robot is in a heavy snowstorm with a snow gust in front of it.
        Most of the points correspond to snowflakes, and no clear object can be identified in the point cloud.
        }
        \label{CBPL_fig:density_field_example}
    \end{figure*}

Finally, one can note that the estimated field also changes in time.
As this additional parameter greatly complexifies the estimation of the field, we assume that the field is constant for a given time interval $[t-\tau/2, t+\tau/2]$ and use all measurements in this interval to estimate the field at time $t$.
We used an interval size of $\tau = \SI{1}{\s}$ in our experiments, and each measurement has an area of $A_c = \SI[parse-numbers=false]{40\times 40}{\cm\squared}$.
Additionally, the field is tessellated into cells of size \SI[parse-numbers=false]{10\times 10}{\cm\squared}.
\autoref{CBPL_fig:density_field_example} shows an example of the resulting density field for different snowstorms.
The left field corresponds to a run of the \ac{CADC} when the snowfall was calm and without wind. 
As a result, the density around the robot is equally distributed and low enough that one can distinguish structural elements in the point cloud.
In the second case, the field is estimated during a heavy gust of wind and snow, yielding to a higher density of snow around the robot.
This event was recorded in the campus of Quebec City, in the same snowstorm represented on \autoref{CBPL_fig:intro}.
Note that the major issue is that because of this higher density, the range of the lidar is greatly reduced, and almost no structured elements are visible in this case.

Then, we demonstrate the close relationship between our method and the concept of visibility that was introduced in the textbook from \citet{Middleton1957} in \citeyear{Middleton1957}, then adapt this concept in the case of lidar sensors.
In the field of meteorology, visibility is defined as the distance at which a black object becomes invisible.
More precisely, an object is visible if the apparent contrast between it and the background is high enough for a human observer to distinguish it.
The contrast $C$ is defined as 
\begin{equation}
  C = \expp{-\sigma r},
  \label{CBPL_eq:apparent_contrast}
\end{equation}
where $r$ is the distance between the observer and the object, and $\sigma$ is the atmospheric extinction coefficient average over the distance $r$ and over the visible spectrum.
The visibility $V$ is then defined as the distance at which the contrast $C$ drops below a threshold $\epsilon$.
In the following, we propose to mimic this metric in the case of lidar measurements instead of a human observer.
We define the notion of p-visibility, which is the distance at which the probability of perceiving an object with a lidar is at most $p$.
For a given field $\lambda(\cdot)$, the probability that the lidar will be able to measure an object without being interrupted by a snowflake is
\begin{equation}
  \begin{aligned}
    \P{\mathtt{object}} &= \expp{-\Lambda(\mathcal{S})} \\
                         &= \expp{-\overline{\lambda} A_\mathcal{S}}, \quad \overline{\lambda}=\frac{\Lambda(\mathcal{S})}{A_\mathcal{S}},
  \end{aligned}
  \label{CBPL_eq:p_measure}
\end{equation}
where $\mathcal{S}$ is the space that the beam needs to cross to reach the object to measure, $A_\mathcal{S}$ its corresponding area and $\overline{\lambda}$ the average density of the field.
Note that this equation is strictly equivalent to \autoref{CBPL_eq:apparent_contrast} with $\sigma=\overline{\lambda}$ and $r=A_\mathcal{S}$.
Assuming the area the beam is traveling in a sector shape, its area is equal to 
$A_\mathcal{S} = \frac12 d^2\alpha,$
where $d$ is the traveled distance of the beam, and $\alpha$ is the aperture angle of the lidar sensor \cite{Laconte2019a}.
Using \autoref{CBPL_eq:p_measure}, we can find the distance $V_p$ at which the probability to perceive the object drops below the probability $p$:
\begin{equation}
  \begin{aligned}
    & \P{\mathtt{object}} = \expp{-\overline{\lambda}\cdot \frac{V_p^2\alpha}{2}} = p \\
    \Leftrightarrow\quad & V_p = \sqrt{\frac{-2\ln{p}}{\overline{\lambda}\alpha}}.
  \end{aligned}
  \label{CBPL_eq:p_visibility}
\end{equation}
Using this metric, one can estimate the p-visibility at any time in a snowstorm.
Furthermore, note that the whole pipeline can easily run online, and thus the p-visibility can be estimated in real time.

One can note strong connections between the contrast visibility in \autoref{CBPL_eq:apparent_contrast} and ours from \autoref{CBPL_eq:p_visibility}. 
Indeed, the atmospheric extinction $\sigma$ corresponds in our case to the local density of the snowfall $\overline{\lambda}$, that both serve the same purpose. 
The higher these coefficients, the less visibility the current observer has.
In the first case, the coefficient depicts the reflectivity of the atmosphere, whereas in our case the reflectivity is replaced by the reflection of the snowflakes.
As such, the density can also be seen as a measure of reflectivity.
The final difference between the equations comes from the fact that a lidar beam travel in a cone shape \cite{Laconte2019a}, contrary to a single light ray, leading to more and more space being occupied by the beam as it travels in space.
Using \autoref{CBPL_eq:p_visibility}, we are able to compute the p-visibility for any given field in real time.
The average density of the field $\overline{\lambda}$ is estimated using data that are at most $\SI{5}{\m}$ away from the robot. 
In the following, we fix the probability at $p=0.5$ and shorten the term 0.5-visibility as simply visibility.
Thus, visibility is the distance at which a lidar beam has half a chance of reaching an object and not being deflected by a snowflake.

As an example, the two density fields depicted in \autoref{CBPL_fig:density_field_example} lead to very different visibilities.
Using a standard lidar sensor, such as the Velodyne HDL-32E or Robosense RS-32 with an aperture angle of \SI{0.085}{\degree}, we can compute the visibility of the lidar for a calm snowfall and a snow gust during a heavier storm. 
For the snowfall (\autoref{CBPL_fig:density_field_example} -- Left), the visibility is equal to approximately \SI{40}{\m}.
In the case of the snow gust (\autoref{CBPL_fig:density_field_example} -- Right), the visibility is reduced to only \SI{5}{\m}, meaning that half of the time the lidar will not be able to measure an obstacle past \SI{5}{\m}, and will instead return noise.

%% file: content/results.tex
\vspace{-1em}%
\section{Results}
\label{CBPL_Sec:Results}
In this section, we evaluate our visibility metric on the \ac{CADC} dataset and correlate it with the performance of a state-of-the-art \index{localization}localization algorithm.
A thorough description of the sensors used to record the \ac{CADC} dataset can be found in~\cite{Pitropov2021}.
Firstly, we describe the algorithm, as well as the localization performance metric that is used throughout this section.
Secondly, we present an analysis of the impact of snowfall on localization.
Then, we discuss the improvements of the localization accuracy that come from filtering the snow from the point cloud.

To evaluate the correlation between our visibility metric and localization accuracy, we have used a lidar-based framework similar to the one described by \citet{Baril2022}.
The framework is based on the \index{Iterative Closest Point}\acf{ICP} registration algorithm. 
This algorithm computes the rigid transformation that minimizes the registration error between two subsequent scans, with the help of a prior based on \ac{IMU} and wheel encoder measurements.
As demonstrated by \citet{Baril2022}, real-time systems cannot compute localization and mapping by taking all lidar points into account due to the high computational complexity. 
As such, a random sub-sampling keeping \SI{70}{\percent} of the input point cloud has been used. 
Additionally, it should be noted that we use a map filter to remove dynamic points after scan registration~\cite{Pomerleau2014}, helping to preserve the consistency of the map.
To evaluate the quality of the localization, we compute the \ac{RPE} as described by \citet{Sturm2012}. 
The estimated trajectory is compared to \ac{GNSS} \ac{RTK} data, considered as ground truth.
We set the window duration to \SI{1}{\s}, as we seek to pinpoint sharp, abrupt localization errors that would lead to dangerous behaviors of the vehicle.
The evaluations have been performed using the \texttt{evo} evaluation library.\footnote{github.com/MichaelGrupp/evo}

\begin{figure}[thbp]
	\includegraphics[width=.95\linewidth]{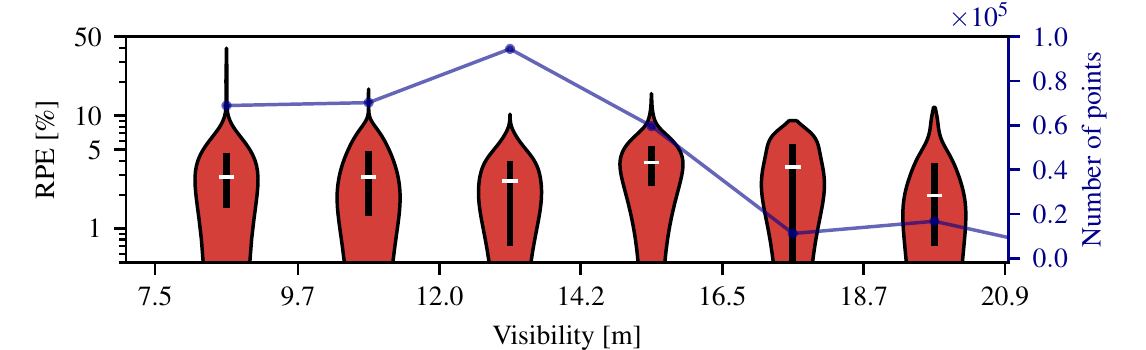}
	\caption{\acf{RPE} correlated with the visibility, estimated as described in \autoref{CBPL_Sec:Theory}. A logarithmic scale is used on the left axis, and data are gathered into bins of \SI{2.2}{\m}. The number of points constituting each bin is displayed on the right axis. As the dataset is focused on harsh weather conditions, most measurements happen in low visibility. No clear correlation between the visibility and the localization error can be established.
	}
	\label{CBPL_fig:rpe_density}
\end{figure}

First, we analyze the impact of snowfall on localization.
\autoref{CBPL_fig:rpe_density} shows the relative pose error as a function of the visibility for all the sequences of the \ac{CADC}.
As we randomly subsample the input point cloud, the dataset have been fully evaluated 50 times, leading to \SI{46}{\hour} of data.
One can see that there is no clear correlation between visibility and localization error.
Surprisingly, the \ac{ICP} algorithm is resilient to a great level of snow precipitation, even when half of the lidar beams cannot see past \SI{8}{\m}.
However, for low visibility, some outliers tends to lengthen the error distribution up to \SI{40}{\percent}.

\begin{figure}[htbp]
	\includegraphics[]{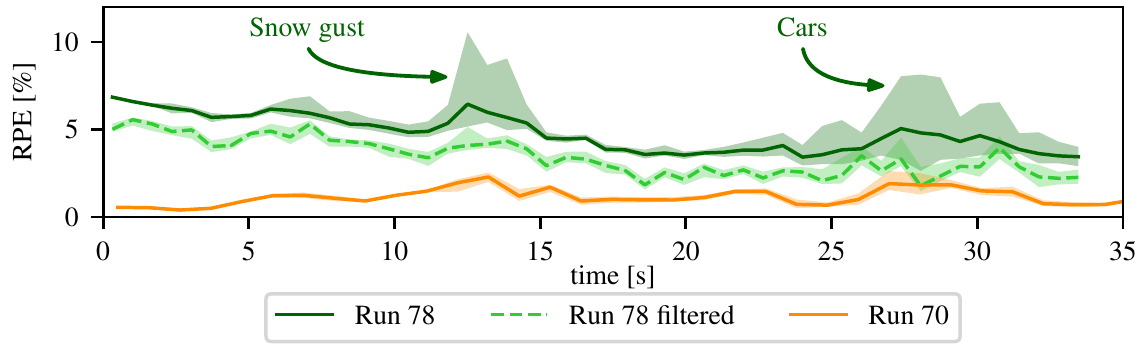}
    \caption{Comparison of the \acf{RPE} of three sequences. 
    Run 70 presents heavy falling snow in an urban environment. 
    Run 78 is recorded on a more rural road than most of the other sequences, and includes events like a snow gust that disrupts the localization. The \ac{DSOR} filter was applied to Run 78 to measure the improvement on localization accuracy.} 
    \label{CBPL_fig:snowgust_time}
\end{figure}

To better highlight these outliers, \autoref{CBPL_fig:snowgust_time} depicts the localization error for two given runs of the \ac{CADC}.
Run 70 contains heavy snow precipitation with an average visibility of ten meters, but maintains a \ac{RPE} below \SI{3}{\percent}.
The environment features several buildings, allowing the \ac{ICP} to rely on many planes to provide an accurate localization. 
Run 78 is recorded in a more rural area, meaning that there are very few landmarks on the side of the road and particularly no buildings.
The environment is therefore much less constrained, resulting in a higher localization error.
In this specific run, the error is relatively constant, but for two distinct events.
The first one corresponds to a heavy gust caused by a passing truck lifting powder snow in front of the lidar sensor.
During such an event, a local, very dense cloud of snowflakes can greatly obstruct the field of view, as depicted in \autoref{CBPL_fig:density_field_example} -- Right.
As the density of snow reaches an extreme level, a large part of the lidar scan consists of beams that hit snowflakes, instead of the static surrounding environment.
The second event simply corresponds to dynamic obstacles (\ie cars) obstructing the field of view of the lidar, leading to a decreased localization accuracy~\cite{Pomerleau2014}. 
One can observe that in the case of wind gusts, the third quartile of the RPE error for this experiment's distribution is very high, going above \SI{10}{\percent}.
Such an error can easily disrupt the entire navigation stack, leading to hazardous behaviors in case of autonomous driving.
This observation leads to theorize that \ac{ICP}-based localization is sensible to which points are randomly filtered in the scan, showing potential localization failures if the system is subject to denser, heterogeneous snowfall.
However, capturing such events is challenging as they are quite rare, and thus more data is needed before being able to draw quantitative evaluations about these events.
We also look at the localization performance gain related to filtering the snow with the \ac{DSOR} filter~\cite{Kurup2021}.
\autoref{CBPL_fig:snowgust_time} shows that an advantage of using a filter is that it greatly reduces the interquartile range of the error during events like a snow gust, meaning that the chance of a total failure of the localization process is also decreased.
Once again, a generalization of this result would require more snow gust data, as these events are scarce and only one snow gust has been recorded in the \ac{CADC}.

\begin{SCfigure}
	\includegraphics[width=.45\linewidth]{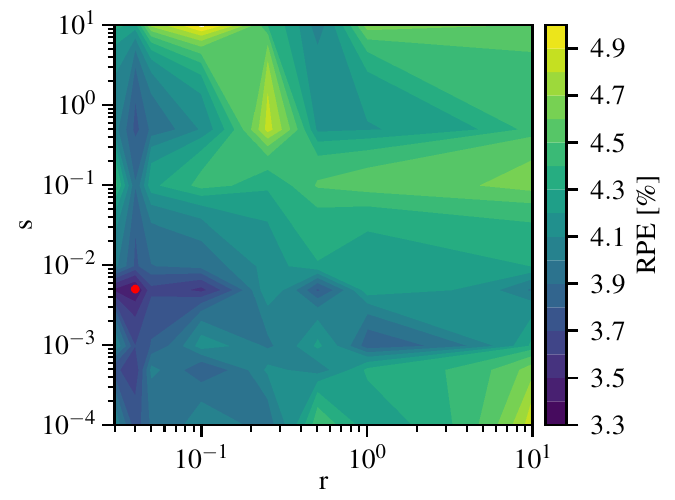}
	\caption{\ac{DSOR} filter applied to Run 78 containing the snow gust event. Multiple couples of parameters ${r,s}$ are tested to find the best configuration, indicated by the red dot, which reduces the median of the \acf{RPE} in the sequence.}
	\label{CBPL_fig:filters}
\end{SCfigure}

\autoref{CBPL_fig:filters} depicts the localization error given different parameters of the \ac{DSOR} filter, which is a state-of-the-art snow removal filter at the time of writing.
The filter has two parameters $s,r \in \SR_{> 0}$.
The parameter $s$ is used in the \ac{SOR} filter~\cite{Rusu2011} to determine the proportion of points that are classified as outliers, while $r$ is added for \ac{DSOR} to scale the filter threshold with the distance between the point and the sensor.
The lower the parameters $s,r$ are, the more aggressive the filter becomes.
Note that setting $s,r\rightarrow +\infty$ is equivalent to not applying any filter on the point cloud.
Overall, we can see that applying the filter at its best configuration yields an improvement of about \SI{1}{\percent} on the median of the \ac{RPE}.
This aggressive tuning explains that the noise of the snow gust can be filtered, even though the points are denser than in usual snow precipitation.

We have evaluated our visibility metric, described in~\autoref{CBPL_Sec:Theory} on all runs of the \ac{CADC} dataset and shown its correlation with localization performance in~\autoref{CBPL_fig:rpe_density}.
Through this result, we observe that extreme precipitation conditions can potentially lead to localization failure, however the data is too sparse to confirm.
We argue that our metric could help standardize the evaluation of the loss of visibility related to precipitation on all datasets.
Our results also motivate the need for more datasets that feature extreme precipitations and visibility loss to challenge state-of-the-art localization algorithms.

%% file: content/conclusion.tex
\section{Conclusion}
\label{CBPL_Sec:Conclusion}
In this chapter, we introduced a novel visibility metric for lidars in snowstorm scenarios, relying on the notion of visibility in the meteorological field. %
We evaluated the \ac{CADC} dataset with this metric, and showed that there is no clear correlation between the visibility in snowstorms and the localization performance of the \ac{ICP} algorithm.
However, events such as snow gusts can lead to localization failure, but available datasets contain too few recordings of these events to quantify their true impact.
As future work, we aim to use our metric to develop precipitation-aware filters and vehicle control.
The visibility metric can also be used to characterize and classify snowstorm datasets from the robotics community. 
Furthermore, research on new datasets targeting specifically extreme and short events, such as snow gusts, would greatly improve the resilience of localization algorithms.

\vspace{-1em}
\section*{Acknowledgements}
This research was supported by the Natural Sciences and Engineering Research Council of Canada (NSERC) through the grant CRDPJ 527642-18 SNOW (Self-driving Navigation Optimized for Winter).

\vskip1.5em